\documentclass[lettersize,journal]{IEEEtran}
\usepackage{amsmath,amsfonts}
\usepackage{algorithmic}
\usepackage{array}
\usepackage[caption=false,font=normalsize,labelfont=sf,textfont=sf]{subfig}
\usepackage{textcomp}
\usepackage{stfloats}
\usepackage{url}
\usepackage{verbatim}
\usepackage{graphicx}
\usepackage[hidelinks]{hyperref}
\usepackage{xcolor}
 
\def\BibTeX{{\rm B\kern-.05em{\sc i\kern-.025em b}\kern-.08em
    T\kern-.1667em\lower.7ex\hbox{E}\kern-.125emX}}
\usepackage{balance}

\usepackage{algorithm}
\usepackage[T1]{fontenc}
\usepackage{booktabs}
\usepackage{multirow}

\newcommand{\g}{^{(g)}}
\newcommand{\x}{\mathbf{x}}

\newcommand{\m}{\mathbf{m}}
\newcommand{\C}{\mathbf{C}}

\DeclareMathOperator*{\argmax}{argmax}

\begin{document}
\title{Solving Deep Reinforcement Learning Tasks with Evolution Strategies and Linear Policy Networks} 
\author{
\IEEEauthorblockA{
    Annie Wong,\textsuperscript{\dag} \and
    Jacob de Nobel,\textsuperscript{\dag} \and
    Thomas Bäck,
    \and
    Aske Plaat, \and
    Anna V. Kononova,
}

\IEEEauthorblockA{Leiden Institute of Advanced Computer Science (LIACS)\\
Leiden University, The Netherlands}
}

\maketitle


\maketitle

\begin{abstract}
Although deep reinforcement learning methods can learn effective policies for challenging problems such as Atari games and robotics tasks, algorithms are complex, and training times are often long. This study investigates how Evolution Strategies perform compared to gradient-based deep reinforcement learning methods. We use Evolution Strategies to optimize the weights of a neural network via neuroevolution, performing direct policy search. 
We benchmark both deep policy networks and networks consisting of a single linear layer from observations to actions for three gradient-based methods, such as Proximal Policy Optimization. These methods are evaluated against three classical Evolution Strategies and Augmented Random Search, which all use linear policy networks. Our results reveal that Evolution Strategies can find effective linear policies for many reinforcement learning benchmark tasks, unlike deep reinforcement learning methods that can only find successful policies using much larger networks, suggesting that current benchmarks are easier to solve than previously assumed. Interestingly, Evolution Strategies also achieve results comparable to gradient-based deep reinforcement learning algorithms for higher-complexity tasks. Furthermore, we find that by directly accessing the memory state of the game, Evolution Strategies can find successful policies in Atari that outperform the policies found by Deep Q-Learning. Evolution Strategies also outperform Augmented Random Search in most benchmarks, demonstrating superior sample efficiency and robustness in training linear policy networks. 
\end{abstract}


\section{Introduction}
\label{sec:introduction}
Gradient-based deep reinforcement learning (DRL) has achieved remarkable success in various domains by enabling agents to learn complex behaviors in challenging environments based on their reward feedback, such as StarCraft~\cite{vinyals2019grandmaster} and Go~\cite{silver2016mastering}. 
However, new methods are often benchmarked on simpler control tasks from OpenAI Gym, including the locomotion tasks from MuJoCo~\cite{haarnoja2018soft} or Atari games~\cite{mnih2015human}. While it simplifies the comparison between different approaches, these benchmarks may lack sufficient complexity, and performance may not always transfer to more complicated tasks. Additionally, several studies have indicated that DRL results are often hard to reproduce~\cite{islam2017reproducibility}, attributing these difficulties to the impact of the random seeds~\cite{henderson2018deep} and the choice of hyperparameters~\cite{eimer2023hyperparameters}. 

Evolution Strategies (ES)~\cite{rechenberg1965,back1991survey}, a family of black-box optimization algorithms from the field of Evolutionary Algorithms (EAs)~\cite{back2023evolutionary}, have been studied as an alternative way to optimize neural network weights, as opposed to conventional gradient-based backpropagation~\cite{salimans2017evolution,such2017deep}. An evolution strategy is used to learn a controller for an RL task by directly optimizing the neural network's weights, which parameterize the RL policy. In this context, the evolution strategy is intrinsically an RL method that performs direct policy search through {\em neuroevolution}~\cite{igel2003neuroevolution}. In supervised learning, gradient-based methods are often much more efficient than ES for training NN weights, though more likely to be trapped in local optima~\cite{mandischer2002comp}. For RL, the need to balance exploration with exploitation in gradient-based approaches incurs more training steps to learn an optimal policy~\cite{igel2003neuroevolution}, making ES an interesting alternative. While EAs are not necessarily more sample-efficient, ES can be more easily parallelized and scaled, offering the possibility for faster convergence in terms of wall-clock time, and, being a global search method, are less likely to get stuck in a local optimum~\cite{morse2016simple}. Additionally, ES do not make any assumptions about the optimization problem, e.g., assuming the environment is Markovian, as long as solutions can be encoded and a fitness function can be defined~\cite{emmerich2018evolution}.

\begin{figure*}[!t]
    \centering
    \includegraphics[trim={0 100pt 0 0},clip,width=0.95\textwidth]{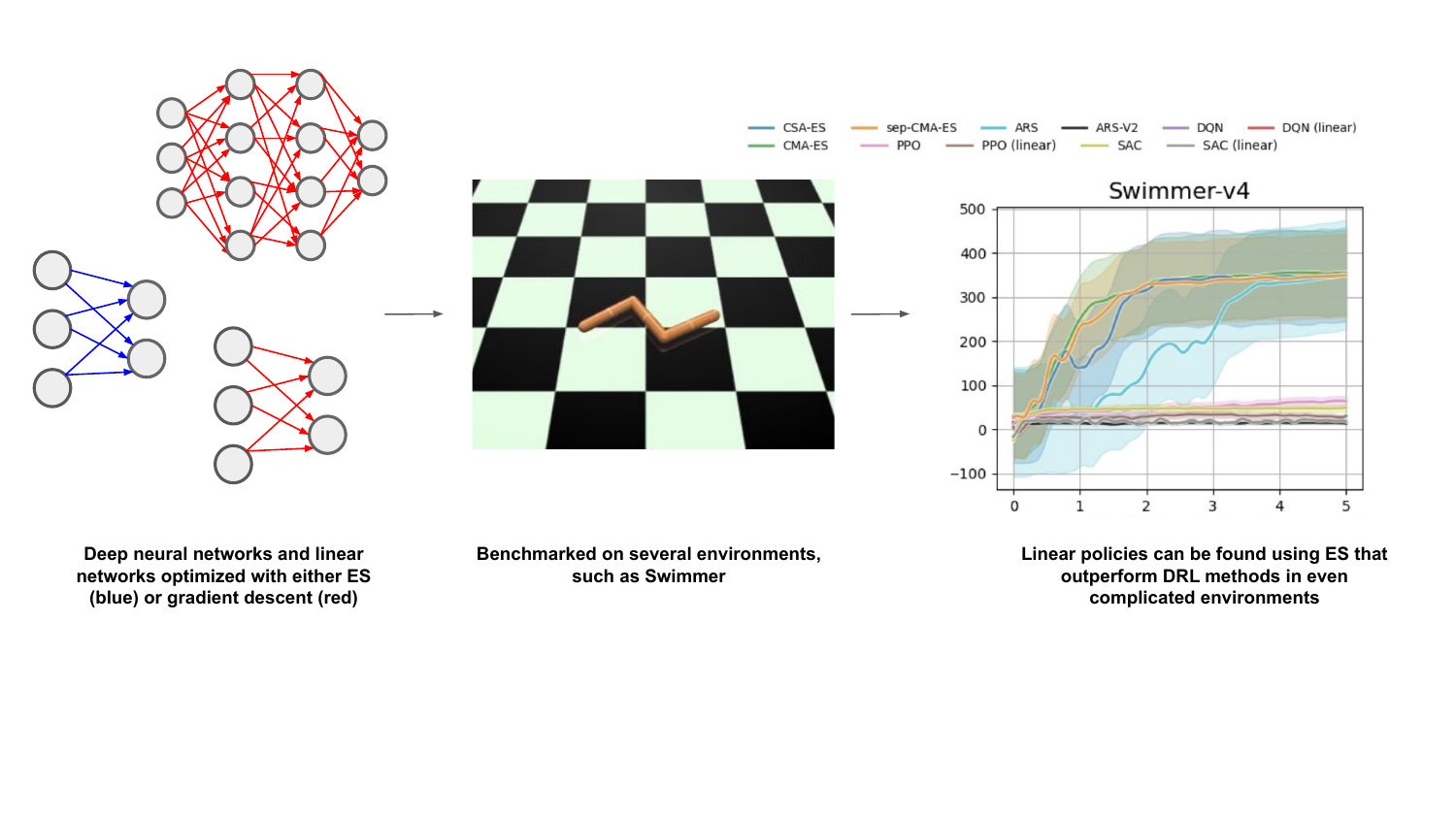}
    \caption{This study investigates how evolution strategies compare to gradient-based reinforcement learning methods in optimizing the weights of linear policies. We use both linear networks as the original DRL architectures to learn policies. We find that ES can learn linear policies for numerous tasks where DRL cannot, and in many instances, even surpasses the performance of the original DRL networks, such as in Swimmer. }
    \label{fig:summary}
\end{figure*}

We benchmark three ES and three gradient-based RL methods on well-known RL tasks to understand the circumstances favoring ES over gradient-based methods. In particular, we study the optimization of policy networks that consist of a single linear layer, from observations to actions for both the ES and gradient-based methods, as low-dimensional controllers of the agent~\cite{ijcai2018p197,rajeswaran2017towards}. Additionally, we include Augmented Random Search (ARS)~\cite{mania2018simple}, which has demonstrated the ability to solve MuJoCo tasks with linear policies as a baseline for comparison. We compare these results to the larger networks used by common gradient-based methods. Our main contributions are as follows:
\begin{itemize}
    \item ES can find effective linear policies for many RL benchmark tasks. In contrast, methods based on gradient descent need vastly larger networks. This finding aligns with previous work demonstrating the potential of simpler policy representations~\cite{mania2018simple,rajeswaran2017towards}. We introduce three advanced ES (CSA-ES, sep-CMA-ES, CMA-ES) that have not been extensively explored in the context of DRL benchmarks.
    \item We find that the ES achieve higher performance compared to ARS across most evaluated games. The ES often converge more quickly to effective linear policies, reducing the overall training time and demonstrating its robustness across tasks and environments.
    \item Contrary to the prevailing view that ES are limited to simpler tasks, they can address more complex challenges in MuJoCo. Gradient-based DRL only performs superiorly in the most challenging MuJoCo environments with more complex network architectures. This suggests that common RL benchmarks may be too simple or that conventional gradient-based solutions may be overly complicated.
    \item Complex gradient-based approaches have dominated DRL. However, ES can be equally effective, are algorithmically simpler, allow smaller network architectures, and are thus easier to implement, understand, and tune (See Figure 1).
    \item We find that advanced self-adaptation techniques in ES are often not required for (single-layer) neuroevolution. 
\end{itemize}

The rest of the paper is organized as follows: Section \ref{sec:backgroundsec} discusses the background and related work of ES and DRL, 
our algorithms are discussed in Section \ref{sec:algorithmssec}, 
the results are in Section \ref{sec:resultssec}, conclusions are in Section \ref{sec:conclusion}.

\section{Background and Related Work}
\label{sec:backgroundsec}

In RL, an agent learns from feedback through rewards and penalties from its environment~\cite{sutton2018reinforcement}. RL problems are formulated as a Markov Decision Process $\langle S, A, P, R, \gamma \rangle$, where $S$ is the set of states in the environment, $A$ the set of actions available to the agent, $P$ the probability of subsequent state transitions, $R$ the reward function, and $\gamma \in [0, 1]$ the discount factor~\cite{bellman1957markovian}. At each time step, the agent is in a state $s_{t} \in S$, takes action $a_{t} \in A$, transitions to $s_{t+1}$, and receives reward $r_{t+1} \in R$. 
A policy $\pi$, parameterized by $\theta$, determines which action to take in each state.
%
%
%
The policy in DRL is typically represented by a deep neural network that maps states to actions (probabilities).
RL aims to find the optimal policy  \( \pi^* \) that maximizes the expected cumulative reward of a state.
RL algorithms approach this goal in different ways~\cite{plaat2022deep}. 
The most common techniques include value-function estimation~\cite{watkins1992q,mnih2015human}, policy gradient methods~\cite{williams1992simple}, actor-critic methods~\cite{konda1999actor,schulman2017proximal,haarnoja2018soft}, and learning a model of the environment~\cite{Hafner2020Dream,plaat2023high}. 

ES are a distinct class of evolutionary algorithms that are particularly suitable for optimization problems in continuous domains. 
ES begin with a population of randomly initialized candidate solutions in \( \mathbb{R}^{n} \), with solutions represented as \( n \)-dimensional vectors denoted by \( \x \) (like the policy)  and a given objective function \( f : \mathbb{R}^n \rightarrow \mathbb{R}\) (like the reward). Via perturbations using a parameterized multivariate normal distribution, selection, and sometimes recombination, solutions evolve towards better regions in the search space~\cite{back1991survey}. Evolving neural networks with EAs is called neuroevolution and can include the optimization of the network's weights, topology, and hyperparameters~\cite{stanley2019designing}. Using ES to evolve a neural network's weights is similar to policy gradient methods in RL, where optimization applies to the policy's parameter space. %

Gradient-based deep RL has successfully tackled high-dimensional problems, such as playing video games with deep neural networks encompassing millions of parameters~\cite{mnih2015human,vinyals2019grandmaster}. 
However, state-of-the-art ES variants are limited to smaller numbers of parameters due to the computational complexity of, for example, adapting the search distribution's covariance matrix.
Covariance Matrix Adaptation Evolution Strategy (CMA-ES) is often used for dimensionality lower than \( n \leq 100 \)~\cite{muller2018challenges}, and problems with a dimensionality \( n \geq 10.000 \) become nearly impossible due to the memory requirements~\cite{loshchilov2014computationally}. 
However, recent advancements have restricted the covariance matrix, in its simplest form, to its diagonal to reduce the computational complexity~\cite{ros2008simple,loshchilov2014computationally,nomura2022fast}. 
Others sample from lower-dimensional subspaces~\cite{maheswaranathan2019guided,choromanski2019complexity}.

In 2015, DRL reached a milestone by achieving superhuman performance in Atari games using raw pixel input~\cite{mnih2015human}. This breakthrough marked a shift in RL towards more complicated, high-dimensional problems and a shift from tabular to deep, gradient-based methods. 
For simpler tasks, the CMA-ES has been used to evolve neural networks for pole-balancing tasks, benefiting from covariance matrix to find parameter dependencies, enabling faster optimization~\cite{igel2003neuroevolution,heidrich2009neuroevolution}. While the use of evolutionary methods for RL can be traced back to the early 90s~\cite{whitley1993,moriarty1996efficient}, the paper by~\cite{salimans2017evolution} rekindled interest in ES from the field of RL as an alternative for gradient-based methods in more complicated tasks. Researchers showed that a natural evolution strategy (NES) can compete with deep RL in robot control in MuJoCo and Atari games due to its ability to scale over parallel workers. In contrast to deep methods where entire gradients are typically shared, the workers only communicate the fitness score and the random seed to generate the stochastic perturbations of the policy parameters. Studies have subsequently demonstrated that simpler methodologies can yield results comparable to NES, such as a classical ES~\cite{ijcai2018p197} and Augmented Random Search (ARS)~\cite{mania2018simple}, which closely resembles a global search heuristic from the 1990s~\cite{salomon1998evolutionary}. In addition, when separating the computer vision task from the actual policy in playing Atari, the size of the neural network can be drastically decreased~\cite{cuccu2018playing}, and policies with a single linear layer, mapping states directly to actions, can effectively solve the continuous control tasks~\cite{ijcai2018p197,rajeswaran2017towards}. The development of dimension-lowering techniques, such as world models~\cite{ha2018world,Hafner2020Dream} and autoencoders~\cite{hinton2006reducing}, also opens up new possibilities for ES to effectively solve more complex problems by simplifying them into more manageable forms.

\section{Methods}
\label{sec:algorithmssec}
We benchmark three ES against three popular gradient-based DRL methods. In addition, we include ARS as a baseline comparison.

\subsection{Gradient-Based Algorithms}
We use three popular gradient-based DRL algorithms: Deep Q-learning~\cite{mnih2015human}, Proximal Policy Optimization~\cite{schulman2017proximal}, and Soft Actor-Critic~\cite{haarnoja2018soft}. We summarize the main gradient-update ideas below.

\subsubsection{Deep Q-Learning}
Deep Q-learning (DQN) combines a deep neural network with Q-learning to learn the value function in a high-dimensional environment~\cite{mnih2015human}. 
Each experience tuple $(s_{t}, a_{t}, r_{t}, s_{t+1})$ is stored in a replay buffer. 
The agent randomly selects a batch of experiences to update the value function. 
The replay buffer breaks the correlation between consecutive experiences, leading to lower variance. 
The primary Q-network weights \( \theta \) are updated every training step by minimizing the expectation of the squared difference between the predicted Q-value of the current state-action pair \( Q(s, a; \theta) \) and the target Q-value \( Q(s', a'; \theta^-) \): 
\[ L(\theta) = \mathbb{E}\left[\left(r + \gamma \max_{a'}Q(s', a'; \theta^-) - Q(s, a; \theta)\right)^2\right] \; \]
The weights from the primary Q-network are copied every $N$ timesteps to a separate target network  \( \theta^- \leftarrow \theta \) to prevent large oscillations in the loss function's target values.

\subsubsection{Proximal Policy Optimization}
Proximal Policy Optimization (PPO) was introduced to improve the complexity of earlier policy gradient methods~\cite{schulman2017proximal}. 
PPO introduces a simpler, clipped objective function:
\[
L^{CLIP}(\theta) = \hat{\mathbb{E}}_t \left[ \min(r_t(\theta) \hat{A}_t, \text{clip}(r_t(\theta), 1 - \epsilon, 1 + \epsilon) \hat{A}_t) \right] 
\]
where $\hat{\mathbb{E}}_t$ denotes the empirical expectation over a finite batch of samples, the probability ratio $r_t(\theta)$ reflects the probability of an action under the current policy compared to the previous policy, $\hat{A}_t$ is the advantage estimate at timestep $t$, and $\epsilon$ is a hyperparameter defining the clipping range. 
The clipping mechanism clips the ratio $r_t(\theta)$ within the range $[1 - \epsilon, 1 + \epsilon]$. 

\subsubsection{Soft Actor-Critic}
SAC objective's function maximizes the expected return and entropy simultaneously to ensure a balanced trade-off between exploitation and exploration: 

\[ \pi^* = \arg\max_\pi \sum_t \mathbb{E}_{(s_t, a_t) \sim \rho_\pi} \left[ r(s_t, a_t) + \alpha H(\pi(\cdot | s_t)) \right] \; , \]
where \( \alpha \) is the temperature parameter that scales the importance of the entropy   \( H(\pi(\cdot | s_t)) \) of the policy \( \pi \) given the state \( s_t \). 
SAC updates its Q-value estimates using a soft Bellman backup operation that includes an entropy term:
\begin{align*}
Q_{\text{new}}(s_t, a_t) &= \mathbb{E}_{s_{t+1} \sim \mathcal{E}} \left[ r(s_t, a_t) \right. \\
&\quad + \left. \gamma \left( Q_{\text{old}}(s_{t+1}, a_{t+1}) - \alpha \log \pi(a_{t+1}|s_{t+1}) \right) \right].
\end{align*}
SAC employs twin Q-networks to mitigate overestimation bias and stabilize policy updates by using the minimum of their Q-value estimates.

\subsection{Evolution Strategies}
\label{sec:es}
ES are designed for solving continuous optimization problems $\text{maximize}_\mathbf{x} f(\mathbf{x})$, where $f: \mathbb{R}^n \to \mathbb{R}$. The ES are used here to optimize the neural network parameters $\theta$ for the policy function $\pi$ through neuroevolution. The objective function to be maximized is the cumulative return over a fixed number of timesteps: $G = \sum_t^T r_t$, calculated over a given episode or rollout with $T$ timesteps. The methods considered here are variants of derandomized ES and use a parameterized normal distribution $\mathcal{N}(\m\g, \sigma\g \C\g)$ to control the direction of the search. The algorithm adapts the parameters of the search distribution to achieve fast convergence (Algorithm \ref{alg:es}).

\begin{algorithm}
\caption{Generic Evolution Strategy}\label{alg:es}
\begin{algorithmic}
\REQUIRE Objective function $f$, number of offspring $\lambda$, number of parents $\mu$, initial estimates for $\m^{(0)}$ and $\sigma^{(0)}$
\STATE $\C^{(0)} \gets \mathbf{I}_n$ 
\FOR {$g$ in $1,2,\dots$} 
    \STATE Sample $\lambda$ candidates $\x_i \sim \mathcal{N}(\m\g, \sigma\g \C\g)$ 
    \STATE Evaluate objective function $f_i \gets f(\x_i)$
    \STATE Select and rank $\mu$ best candidates
    \STATE Adapt $\m^{(g+1)}, \sigma^{(g+1)}, \C^{(g+1)}$
\ENDFOR
\end{algorithmic}
\end{algorithm}

At each iteration, the evolution strategy samples $\lambda$ offspring from its mutation distribution. By selecting the $\mu \leq \lambda$ most promising offspring to update its parameters, it moves to regions of higher optimality. After sorting the $\mu$ offspring by fitness ranking, the mean of the search distribution is updated via weighted recombination: 
\begin{equation*}
    \m^{(g+1)} = \m\g + c_m \sum_{i=1}^{\mu} w_i (\x_{i} - \m\g)
\end{equation*}

\begin{figure}[!t]
    \centering
    \includegraphics[width=\linewidth]{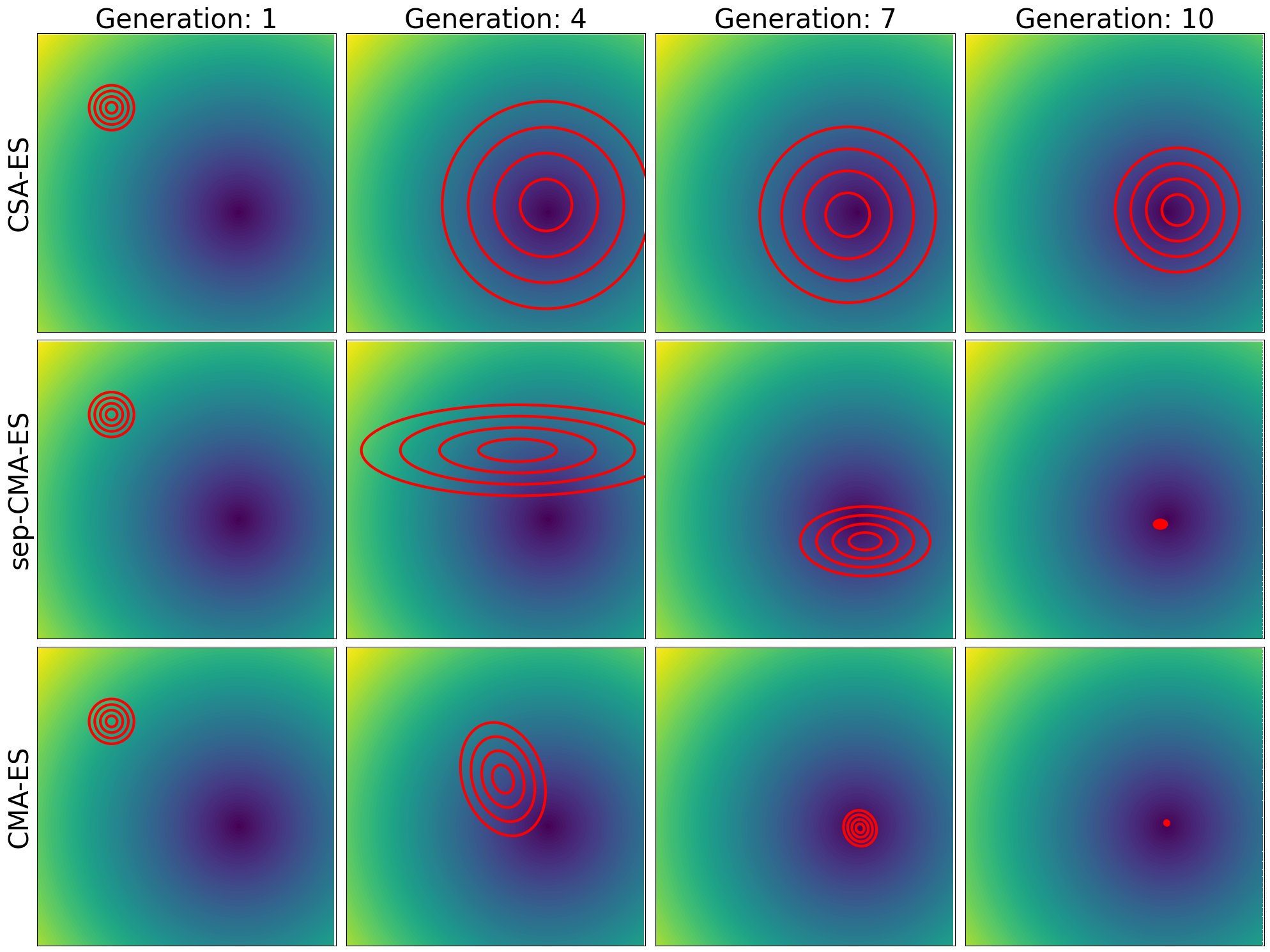}
    \caption{Adaptation of the mutation distribution for three different Evolution Strategies for the first ten generations of a two-dimensional quadratic function. Function values are shown with color; darker indicates lower (better). Top row: mutation distribution for CSA-ES; middle row: sep-CMA-ES; bottom row: CMA-ES }
    \label{fig:adaptation}
\end{figure}

The ES adapt, with increasing complexity, the scale $\sigma^{(g)}$ and shape $\C\g$ of the mutation distribution. The Cumulative Step-size Adaptation Evolution Strategy (CSA-ES) only adapts $\sigma^{(g)}$, producing exclusively isotropic (i.e. $\C\g = \mathbf{I}_n$) mutations during optimization. The separable Covariance Matrix Adaptation Evolution Strategy (sep-CMA-ES) additionally adapts the diagonal entries of the covariance matrix $\C\g$, producing mutation steps with arbitrary scaling parallel to each coordinate axis. Finally, the CMA-ES adapts the full covariance matrix, which allows the mutation distribution to scale to arbitrary rotations. Figure \ref{fig:adaptation} illustrates the evolution of the mutation distribution for each of these three methods when optimizing a two-dimensional quadratic function. The figure shows that the mutation distribution guides the search, favoring selected mutation steps with high probability~\cite{hansen2001completely}. The CSA-ES uses a process called cumulation of historically selected mutation steps to update the value of the global step size parameter $\sigma^{(g)}$. We implemented the algorithm following ~\cite{chotard2012cumulative}, using recommended hyperparameter settings. While several modifications of the CMA-ES have been developed over the years, we implemented a canonical version of the algorithm, as first introduced in~\cite{hansen2001completely}. The update of the full covariance matrix becomes computationally impractical for $n > 100$, but the sep-CMA-ES, which we implemented according to~\cite{ros2008simple}, does not suffer from this restriction. As shown in Figure~\ref{fig:adaptation}, this algorithm only computes variances for each coordinate axis, which makes it applicable to much higher dimensions, as the computational complexity for the update of the mutation distribution scales only linearly with $n$.

\subsection{Augmented Random Search}
Augmented Random Search (ARS) is another gradient-free method~\cite{mania2018simple} that directly optimizes the policy parameters $\theta$ by maximizing the cumulative episodic return, similar to the ES introduced in the previous Section (\ref{sec:es}). Although its name might suggest otherwise, the method is very similar to ES and closely resembles evolutionary gradient search without self-adapation~\cite{salomon1998evolutionary}. The method demonstrated the capability of finding effective linear policies for several MuJoCo benchmarks~\cite{todorov2012mujoco} and is considered in standard RL baselines~\cite{raffin2021stable}. In short, the method samples a population of perturbations $\delta_i \sim \mathcal{N}(0, \mathbf{I})$ at every iteration. It calculates a gradient estimate with respect to $\delta_i$ to update $\theta$ via a weighted average of the observed cumulative reward. The update is scaled by the observed cumulative reward's standard deviations $\sigma_R$ to improve learning stability. Like ES, ~\cite{mania2018simple} identifies that using a smaller subset of top-performing individuals instead of the entire population to do the parameter update positively impacts performance. There are two versions of ARS, which we label ARS-V1 and ARS-V2 respectively. V2 extends on V1 by calculating a rolling mean and variance to standardize the states to zero mean and unit variance. 
 
\begin{figure*}[!t]
    \centering
    \includegraphics[width=\linewidth]{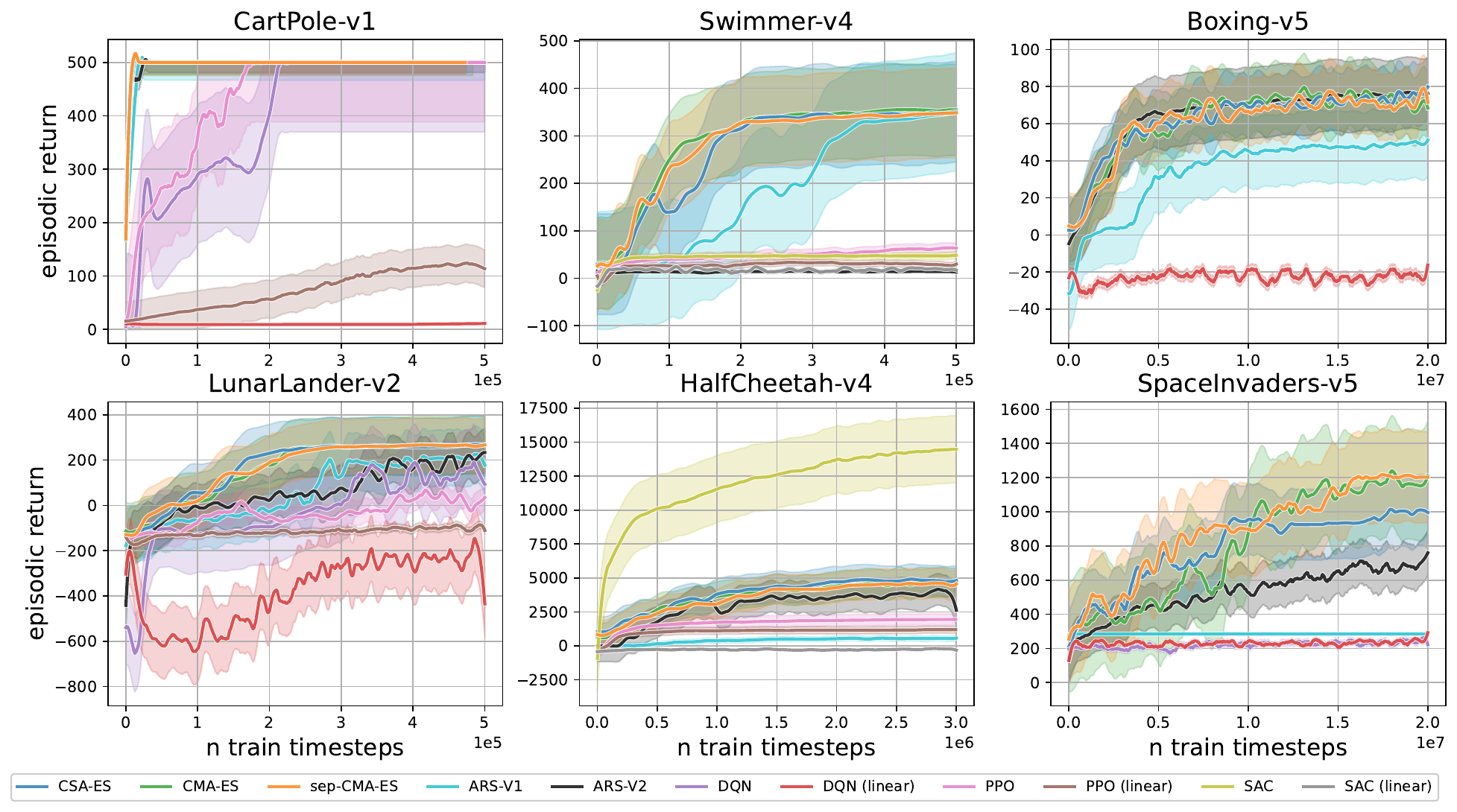}
    \caption{Training curves for the CartPole, LunarLander, Swimmer, HalfCheetah, Boxing, and SpaceInvaders environments. Episodic return (calculated using 5 test episodes) vs. the number of training timesteps is shown. Each curve represents the median of 5 trial runs conducted with different random seeds; the shaded area denotes standard deviations. The results show that the ES solve the classic control environments Cartpole and LunarLander almost immediately. ARS takes slightly longer but outperforms the gradient-based methods. Even for the more difficult Swimmer environment, ES and ARS find a linear policy outperforming DRL in terms of timesteps and performance. While SAC outperforms all other methods in Cheetah, linear ES outperforms classic PPO. For the Atari environments, Boxing and Space Invaders, ES is able to learn a linear policy from the RAM input, while linear DQN fails to do so. Only for Boxing does DQN find a successful policy. ARS is able to improve on a policy for Boxing, although it does not perform as well as ES. However, for Space Invaders, ARS fails to learn a policy.}
    \label{fig:combined6}
\end{figure*}

\subsection{Network Architecture}
We compare the performance of linear policies trained through neuroevolution by ES or ARS with gradient-based methods inspired by earlier studies demonstrating this approach's feasibility~\cite{mania2018simple,rajeswaran2017towards}. For the ES and ARS, only linear policies are trained, defined as a linear mapping from states to actions, activated by either an $\argmax$ or $\tanh$ function for discrete and continuous action spaces, respectively (no hidden layer: a fully connected shallow network). We use the gradient-based methods to train the same linear policies for each control task. Table \ref{tab:weights} (Appendix) shows the number of trainable weights for each environment for a linear policy. Additionally, a network architecture based on the original studies for each gradient-based method is trained for comparison. We employ the architecture from the original studies for PPO~\cite{schulman2017proximal} and SAC~\cite{haarnoja2018soft}. For DQN, we use the default architecture from CleanRL's library, which has been tested across multiple control environments and showed good results~\cite{huang2022cleanrl}. Specifics for these architectures and other hyperparameters can be found in the Appendix. We do not train these deep architectures using ES and ARS, as they only serve as a benchmark to demonstrate the intended usage of the gradient-based algorithms, and self-adaptation mechanisms are increasingly less useful for such high dimensions~\cite{ijcai2018p197}. 

\begin{table*}[!t]
\centering
\caption{Average maximum score per game across trials. The rightmost column shows the best-performing ES episode per game. For comparison, the scores for DQN, a random agent, and a Human player taken from \protect\cite{mnih2015human} are shown. The highest average scores are shown in boldface. The results show that ES achieved the highest scores in two games: sep-CMA attained the highest score in Atlantis and CMA-ES in Boxing, outperforming the Human player and DQN. For the other games, the highest score is attained by the DQN agent, although CMA-ES achieves a score almost identical to DQN on SpaceInvaders. Furthermore, the best ES policy often matches the performance of DQN, demonstrating that a linear policy can be equally effective.}
\scriptsize
\begin{tabular}{c|ccc|c c| ccc|c}
Algorithm & CSA-ES & CMA-ES & sep-CMA & ARS-V1 & ARS-V2 & Random & Human & DQN & ES* \\ 
     \cmidrule(lr){2-2} \cmidrule(lr){3-3} \cmidrule(lr){4-4} \cmidrule(lr){5-5} \cmidrule(lr){6-6} \cmidrule(lr){9-9}
& M(SD) & M(SD) & M(SD) &  M(SD) &  M(SD) &   &  &   M(SD) \\ 
\hline 
Atlantis & 84690 (1.3$\cdot10^4$) & 87100 (1.1$\cdot10^4$) & \textbf{88580} (9.1$\cdot10^3$) & 53880 (3.5$\cdot10^3$) & 61930 (5.7$\cdot10^3$) & 12850 & 29028 &  85641 (1.7$\cdot10^4$) &  103500 \\
B. Rider &2215 (1088) & 1967 (582) & 2222 (721) & 1190 (216) & 1152 (264) & 363.9 & 5775 &  \textbf{6846} (1619) & 5072 \\
Boxing & 96.0 (3.8) & \textbf{96.8} (3.2) & 95.1 (4.3) & 62.6 (4.9) & 83.2 (1.8) & 0.1 & 4.3 &  71.8 (8.4) & 100 \\
C. Climber & 36170 (1.0$\cdot10^4$) & 29290 (6.5$\cdot10^3$) & 32940 (7.8$\cdot10^3$) & 8020 (3.2$\cdot10^3$) & 20070 (5.5$\cdot10^3$)  & 10781 & 35411 & \textbf{114103} (2.2$\cdot10^4$) & 57600 \\
Enduro & 65.1 (22.1) & 58.9 (17.8) & 69.0 (22.9) & 104.2 (32.1) & 83.5 (33.5) & 0 & 309.6 &  \textbf{301.8} (24.6) & 102 \\
Pong & 5.7 (4.0) & 7.4 (10.3) & 7.1 (9.4) & -13.9  (2.2) & -11.5 (2.4) & -20.7 & 9.3 &  \textbf{18.9} (1.3) & 21 \\
Q*Bert & 7355 (4037) & 5732 (2339) & 7385 (3384) & 760 (324) & 3390  (2.2$\cdot10^3$) & 163.9 &  13455 &  \textbf{10596} (3294) & 14700 \\
Seaquest & 959 (204) & 948 (117) & 954 (143) & 526 (199) & 814 (19) & 68.4 & 20182  &  5286 (1310) & 1470\\
S. Invaders & 1640 (567) & 1972 (332) & 1488 (191) & 432 (214) & 914 (210) & 148 & 1652 &  \textbf{1976} (893) & 2635 \\
\hline 
\end{tabular}
\label{tab:atari}
\end{table*}

\begin{table*}[!h]
\centering
\scriptsize
\caption{Maximum episodic return for each environment averaged over five trials; each trial uses the number of training timesteps specified in the table. Linear networks are marked with a * symbol. The table shows that ES and DRL vary in effectiveness across tasks. For example, in simpler tasks like CartPole, ES all achieve a maximum score of 500, matching the performance of classic DQN and PPO. However, in the more complex MujoCo tasks, classic gradient-based methods, particularly SAC, outperform ES. Despite this, linear ES algorithms consistently outperform their linear gradient-based counterparts across various environments, suggesting that gradient-based algorithms may be less effective at discovering linear strategies than ES algorithms.} 
\begin{tabular}{l|c|ccc|cc|ccc|ccc}
 & Timesteps & CSA-ES & CMA-ES & sep-CMA-ES & ARS-V1 &  ARS-V2 & DQN & PPO & SAC & DQN* & PPO* & SAC* \\  
 \hline 
CartPole-v1 & $5\cdot 10^5$ & \textbf{500} & \textbf{500} & \textbf{500} & \textbf{500} & \textbf{500} & \textbf{500} & \textbf{500} &-  & 17 & 197 &  -  \\
Acrobot-v1 & $5\cdot 10^5$ & -75	& -73 &-75 &	-69&	-69&	\textbf{-65}&	-69 & - &-76&	-133& - \\
Pendulum-v1 & $5\cdot 10^5$ & -669 & -668 & -731 & -898 & -964 &  -  & -805 & \textbf{-125} &  -  & -955 & -1063 \\ 
 \hline 
LunarLander-v2 & $5\cdot 10^5$ & 268 & \textbf{273} & 269 & 204 & 263 & 242 & 145 &  -  & -18 & -37 &  -  \\ 
BipedalWalker-v3 & $2\cdot 10^6$ & 228 & 237 & 190 & 3 & 104 &  -  & \textbf{278} & - &  -  & 214 & - \\ 
 \hline 
Swimmer-v4 & $5\cdot 10^5$ & 281 & 315 & 273 & \textbf{345} & 15 & -  & 62 & 50 &  -  & 33 & 30 \\ 
HalfCheetah-v4 & $3\cdot 10^6$ & 4873 & 4705 & 4623 & 677 & 4296 &  -  & 2566 & \textbf{14662} &  -  & 2254 & 150 \\ 
Hopper-v4 & $1\cdot 10^6$ & 2594 & 2913 & 2721 & 1128 & 2292 & -  & 2938 & \textbf{3620} &  -  & 910 & 138 \\ 
Walker2d-v4 & $2\cdot 10^6$ & 1924 & 2359 & 2545 & 1296 & 2288 & -  & \textbf{4003} & 3790 &  -  & 490 & 766 \\ 
Ant-v4 & $1\cdot 10^7$ & 2647 & 2819 & 2684 & 994 & 2457 &-  & 2904 & \textbf{6372} &  -  & 4467 & -28 \\ 
Humanoid-v4 & $1\cdot 10^7$ & 821 &  -  & 830 &955  & 1574 & -  & 2490 & \textbf{7282} &  -  & 3561 & 652 \\ 
\hline 
\end{tabular}
\label{tab:stats}
\end{table*}

\subsection{Experimental Setup}
We conduct experiments on common control tasks of varying complexity levels from the Gymnasium API~\cite{towers_gymnasium_2023}. For each of the considered environments, five runs using different random seeds are conducted for each algorithm/control task combination to test the stability of each approach. Value-based DQN is only used for environments with discrete action spaces, SAC for continuous action spaces, and PPO, ES, and ARS are used for both action spaces. The RL algorithms are implemented using the cleanRL library\footnote{\url{https://github.com/vwxyzjn/cleanrl}} that has been benchmarked across several environments; we removed the hidden layers for the linear network. For ARS, we evaluate both ARS-v1 and ARS-v2 (state normalization) with the enhancement of using top-performing directions and use the implementation of the Stable Baselines3 library. 
The specifics of the implementations are detailed in the code repository accompanying this paper \footnote{https://github.com/ann-w/solving\_drl\_tasks\_with\_es\_and\_linear\_policy\_networks}.


Since the environments are stochastic, we report the median episodic return, calculated over five test episodes. As was discussed in~\cite{salimans2017evolution}, for ES, the wall-clock time required to solve a given control task decreases linearly with the number of parallel workers available. This allows us to perform substantially more evaluations of the environment than is feasible with gradient-based RL. For fairness of comparison, we limit the difference in the number of training time steps allowed by a single order of magnitude. Specific hyper-parameters used for each environment, including hardware, can be found in the Appendix. For the ES, we initialize each experiment with $\m^{(0)} = \mathbf{0}$. We calculate a rolling mean and variance of the observations of the environment during each run. These values are then used to normalize each state observation to standard normal entries by subtracting this rolling mean and dividing by the standard deviation. 

\subsubsection{Classic RL Environments}
The first set of experiments includes the classic control tasks Cartpole, Acrobot, and Pendulum. We include BipedalWalker and LunarLander from the Box2D simulations for slightly more complex dynamics. Each run uses a maximum of 500\,000 timesteps for each environment. The exception is the BipedalWalker task, for which $2\cdot10^6$ timesteps are used.

\subsubsection{MuJoCo Simulated Robotics}
We evaluate the algorithms on the MuJoCo environments~\cite{todorov2012mujoco} for higher complexity levels, including Hopper, Walker2D, HalfCheetah, Ant, Swimmer, and Humanoid. Table ~\ref{tab:hyper} (Appendix) provides training details. As was noted by~\cite{mania2018simple}, ES have exploration at the policy level, whereas gradient-based methods explore on the action level. In the locomotion tasks, a positive reward is provided for each time step where the agent does not fall over. This causes the ES method to stay in a local optimum when the agent stays upright but does not move forward (the gradient-based methods do not get stuck). Following~\cite{mania2018simple}, we modified the reward function for these environments for ES, subtracting the positive stability bonus and only rewarding forward locomotion.

\subsubsection{Atari Learning Environment}
Finally, we benchmark DQN against the ES with linear policies on games from the Atari suite. To demonstrate the effectiveness of linear policies for these high-dimensional tasks, we take inspiration from the approach by~\cite{cuccu2018playing} and separate the computer vision task from the control task. We train the ES agents on the 128 bytes of the Random Access Memory (RAM) in the simulated Atari console. This drastically reduces the input dimensionality of the controller, allowing for the training of smaller policies. This assumes that the random access memory sufficiently encodes the state of each game without having to extract the state from the raw pixel images.
It should be noted that not for all games is RAM information sufficient to train a controller and that for some games, DQN is easier to train on pixel images than on RAM input~\cite{sygnowski2017learning}. Since we evaluate linear policies, we fix frame skipping to 4, with no sticky actions~\cite{machado2018revisiting}, similar to the settings used in~\cite{mnih2015human}. For each run, the ES were trained for 20\,000 episodes, where the maximum episode length was capped at 270\,000 timesteps. The gradient-based methods were trained for a maximum of $2\cdot10^7$ timesteps. 

\section{Results}
\label{sec:resultssec}
In this section, we present the results of our experiments (more details, including training curves and tables with summary statistics, can be found in the Appendix). Here, we focus on a selection of environments (Figure \ref{fig:combined6}).

\begin{table*}[!h]
\centering
\scriptsize
\caption{Lowest number of timesteps required to reach reward threshold in any of the five trial runs. The  $\infty$ symbol indicates that the threshold was not attained, and the '-' symbol indicates no available data. In less complex tasks such as CartPole and Acrobot, ES consistently attain the threshold reward more quickly than DRL, requiring only a few thousand time steps compared to the tens of thousands needed by gradient-based methods. In more complex tasks like HalfCheetah and Ant, DRL methods, especially SAC, tend to reach the threshold more efficiently, with SAC achieving it in just 50\,000 timesteps for HalfCheetah and 400\,000 for Ant, compared to millions of timesteps needed by ES algorithms. It is important to note that although ES require more timesteps than DRL methods, this doesn't necessarily translate to longer wallclock time, as ES can be effectively parallelized. In the Swimmer task, ES achieves the threshold in 300.000 to 700.000 timesteps, while DRL methods fail to reach the threshold.}

\begin{tabular}{c|c|ccc|cc|ccc|ccc}
 & Threshold & CSA-ES & CMA-ES & sep-CMA-ES & ARS-V1 & ARS-V2 & DQN & PPO & SAC & DQN* & PPO* & SAC* \\ 
 \hline 
CartPole-v1 & 475 & $3\cdot 10^3$ & $2\cdot 10^3$ & $3\cdot 10^3$ & $1\cdot 10^3$ & \textbf{560} & $2\cdot 10^4$ & $6\cdot 10^4$ &  -  & $\infty$ & $\infty$ &  -  \\ 
Acrobot-v1 & -100 & $\mathbf{4\cdot 10^3}$ & $5\cdot 10^3$ & $\mathbf{4\cdot 10^3}$ & $7\cdot 10^3$ & $1\cdot 10^4$ & $2\cdot 10^4$ & $7\cdot 10^4$ &  -  & $1\cdot 10^5$ & $\infty$ &  -  \\ 
Pendulum-v1 & -100 & $\infty$ & $\infty$ & $\infty$ & $\infty$ & $\infty$ & -  & $\infty$ & $\infty$ &  -  & $\infty$ & $\infty$ \\ 
\hline 
LunarLander-v2 & 200 & $\mathbf{5\cdot 10^4}$ & $7\cdot 10^4$ & $6\cdot 10^4$ & $2\cdot 10^5$ & $2\cdot 10^5$ & $3\cdot 10^5$ & $4\cdot 10^5$ &  -  & $\infty$ & $\infty$ &  -  \\ 
BipedalWalker-v3 & 300 & $\mathbf{2\cdot 10^6}$ & $\mathbf{2\cdot 10^6}$ & $5\cdot 10^6$ & $\infty$ &$\infty$ &  -  & $\infty$ & - &  -  & $\infty$ & - \\ 
\hline 
Swimmer-v4 & 360 & $4\cdot 10^5$ & $\mathbf{3\cdot 10^5}$ & $7\cdot 10^5$ & $\infty$ & $\infty$ &  -  & $\infty$ & $\infty$ &  -  & $\infty$ & $\infty$ \\ 
HalfCheetah-v4 & 4800 & $2\cdot 10^6$ & $1\cdot 10^6$ & $2\cdot 10^6$ & $\infty$ & $2\cdot 10^6$&  -  & $\infty$ & $\mathbf{5\cdot 10^4}$ &  -  & $\infty$ & $\infty$ \\ 
Hopper-v4 & 3000 & $4\cdot 10^5$ & $3\cdot 10^5$ & $1\cdot 10^6$ & $\infty$ & $1\cdot 10^6$&  -  & $3\cdot 10^5$ & $\mathbf{1\cdot 10^5}$ &  -  & $\infty$ & $\infty$ \\ 
Walker2d-v4 & 3000 & $6\cdot 10^6$ & $9\cdot 10^5$ & $1\cdot 10^6$ & $\infty$& $2\cdot 10^6$&  -  & $\mathbf{7\cdot 10^5}$ & $1\cdot 10^5$ &  -  & $\infty$ & $\infty$ \\ 
Ant-v4 & 5000 & $3\cdot 10^7$ & $3\cdot 10^7$ & $2\cdot 10^7$ & $\infty$ & $\infty$& -  & $\infty$ & $\mathbf{4\cdot 10^5}$ &  -  & $\infty$ & $\infty$ \\ 
Humanoid-v4 & 6000 & $\infty$ &  -  & $5\cdot 10^7$ & $\infty$ & $\infty$&  -  & $\infty$ & $\mathbf{1\cdot 10^6}$ &  -  & $\infty$ & $\infty$ \\ 
\hline 
\end{tabular}
\label{tab:stats2}
\end{table*}

\subsection{Classical RL Environments}
The first column of Figure \ref{fig:combined6} shows the training curves 
on the CartPole 
and LunarLander environments (See Appendix for more results). The ES outperform the deep gradient-based methods for both environments with just a simple linear policy. For CartPole, the results are even more surprising, as the ES policies can solve the environment in the first few iterations of training through pure random sampling from a standard normal distribution. This observation also holds for ARS-v1 and ARS-v2. The deep gradient-based methods, on the other hand, require around $2\cdot10^5$ timesteps to solve CartPole. The gradient-based methods are unable to find a good linear policy for CartPole. This pattern persists for LunarLander, where, even though the ES requires around $2\cdot10^5$ timesteps to solve the environment, the gradient-based methods cannot find a good linear policy at all. While the deep gradient-based methods eventually seem to catch up to the ES, it still requires more than $5\cdot10^5$ timesteps to train a stable policy for LunarLander. ARS-v1 and ARS-v2 perform better than the gradient-based methods but are less sample-efficient than ES, needing around $4 \cdot 10^5 $ timesteps to achieve a reward of 200. For the BipedalWalker task (see Appendix), the ES, SAC, and PPO can find good policies, with the gradient-based methods requiring fewer timesteps. ARS-v1 fails to learn a good policy for this environment. By applying state normalization, ARS-v2 performs better but not as well as the other methods. Only SAC comes close to solving the Pendulum environment within $5\cdot10^5$ timesteps. For Acrobot, again, the ES and ARS can solve the environment almost instantly, while the gradient-based methods require a good number of environment interactions to do so. 

\subsection{MuJoCo Simulated Robotics}
The center column of Figure \ref{fig:combined6} shows that the ES policies are much better at finding a policy for the Swimmer environment compared to ARS and the gradient-based methods. Surprisingly, ARS-v2 cannot learn an effective policy while ARS-V1 is able to, albeit slower than the ES. At the same time, for HalfCheetah, SAC greatly outperforms all other methods. Still, ES outperform PPO and ARS. Moreover, none of the gradient-based methods can find a good linear policy for HalfCheetah and Swimmer. This pattern holds for almost all MuJoCo experiments; PPO can only find a linear policy with decent performance in the Ant environment. Overall, as the number of weights increases, as seen in tasks like Hopper, Walker2d, and Humanoid, the performance of ES and ARS fall behind that of deep gradient-based methods (see Table \ref{tab:weights} in the Appendix). Nevertheless, even though ES generally requires more timesteps, they can still find good linear policies for most environments, which are just as effective as policies found by vastly larger networks (see Table ~\ref{tab:stats} in the Appendix). This is in line with findings in the ARS study, which also demonstrated the effectiveness of linear policies in the MuJoCo environments~\cite{mania2018simple}. Note that~\cite{mania2018simple} considers more environment interactions for these tasks. Even for the most complex of these environments, Humanoid, the ES are able, in several trials, to find a linear policy that has a higher episodic return, $\approx 8000$ (averaged over 5 test episodes with different random seeds), than was found by the best deep gradient-based method, SAC. 
Furthermore, ES timesteps are quicker and easier to parallelize~\cite{salimans2017evolution}, meaning experiments can take a considerably shorter amount of runtime in practice.

\subsection{Atari Learning Environment}
The last column of Figure \ref{fig:combined6} shows the training curves of ES, ARS, and DQN for the Atari games SpaceInvaders and Boxing. The figure shows that when training agents that use the controller's RAM state as observations, ES and ARS outperform linear DQN in most cases. CrazyClimber (Appendix) is the only exception for which ES and linear DQN performance is similar. ARS-V1 achieves the highest score in two games, namely Enduro and Beamrider. Even comparing against deep DQN trained on RAM memory, we find that for both the games in Figure \ref{fig:combined6}, the ES yields better policies and requires fewer environment interactions. In addition, Table \ref{tab:atari} shows the average highest score per trial for each of the tested games for the ES, compared against the numerical results presented in~\cite{mnih2015human} for a Human, Random, and a DQN player that uses pixel input. The table additionally shows the highest score attained by any ES in any trial, averaged over 5 test episodes. For both Atlantis and Boxing, an ES achieves the highest score. For all the other games tested, the DQN agent earned a higher score than all RAM-trained ES, although CMA-ES achieved a score almost identical to DQN on SpaceInvaders. This score is attained by an agent that uses a linear policy consisting of only 768 weights, while the policy trained by DQN has $\approx$ 1.5$\cdot10^6$ (pixel-based, deep policy network). Moreover, the best-found policy by any ES is often competitive with DQN. This indicates that a linear policy, which is competitive with pixel-based DQN, does exist.

\section{Discussion and Conclusion}
\label{sec:conclusion}
In this study, we have explored different ways to optimize reinforcement learning policies with conventional deep learning gradient-based backpropagation methods as used in DQN, PPO, and SAC, as well as with three evolution strategy methods and ARS. We have applied these methods to several classic reinforcement learning benchmarks. We trained the regular deep network as conventionally used for these methods and a neural network with no hidden layers, i.e., a linear mapping from states to actions, as a low-complexity controller for each environment. In many tested environments, the linear policies trained with the ES are on par or, in some cases, even better controllers than the deep policy networks trained with the gradient-based methods. In addition, ES outperforms ARS in most environments. Linear ES converges more quickly to effective policies, reducing the overall training time and demonstrating robustness across tasks and environments. The gradient-based methods are often ineffective at training simple policies, requiring much deeper networks. For our experiments on Atari, we find that by accessing the RAM memory of the Atari controller, ES methods can find a linear policy that is competitive with "superhuman" DQN~\cite{mnih2015human}. It should be noted that there are certain high-complexity environments where the deep gradient-based methods yield better policies, e.g.\ SAC for HalfCheetah. However, even for these environments, linear policies exist that are competitive and much more easily interpretable. 

We conclude that conventional gradient-based methods might be overly complicated or that more complex benchmarks are required to properly evaluate algorithms. In fact, even for our experiments' most complex locomotion task, the Humanoid environment, the CMA-ES found a linear policy that was competitive with state-of-the-art methods. As the ES are stochastic algorithms, they could not find these policies for every trial run, but our results show that such policies do exist. We expect the search landscapes for these environments to be deceptive and multimodal, and future work could help discover effective algorithms for more consistently training these linear policy networks, for example, using niching methods~\cite{shir2005niching}.  
We hypothesize that gradient-based methods may struggle to find linear policies due to the multimodal nature of the search landscape, a phenomenon also seen in supervised learning~\cite{kawaguchi2016deep}. Counterintuitively, with gradient-based methods, it seems more straightforward to train deeper architectures than shallower ones with far fewer weights, as shown by~\cite{schwarzer2023bigger}. We note that gradient-based methods are essentially local search methods, requiring heuristically controlled exploration, while ES, in the early phases of optimization, are performing global search, producing more diverse solutions. This also becomes evident in simpler environments, such as CartPole, where the ES can almost instantly sample the optimal policy, while the gradient-based methods have a much harder time. 

Moreover, we find many counterexamples to the prevailing view that ES are less sample efficient than the gradient-based methods. For many low to medium-complexity environments, the ES are more sample efficient and require fewer environment interactions than the deep gradient-based methods. On the other hand, for the more complex environments, and with increasing dimensionality, we find that the ES can take more time steps to converge than the deep gradient-based methods. This is to be expected, as the self-adaptation mechanisms central to the ES become increasingly ineffective for larger dimensions~\cite{ijcai2018p197,muller2018challenges}. 
We have compared three ES that, with increasing levels of complexity, adapt the shape of the mutation distribution to converge the search. Our results indicate that updates of the covariance matrix are often not required and that performing step size adaptation is sufficient. While we expect the search landscape to be multimodal, relative scaling and rotation of search coordinates seem absent, allowing isotropic mutations to be effective for these problems. This would also explain the effectiveness of the approach demonstrated in~\cite{maheswaranathan2019guided}, which would be heavily impacted by conditioning on the search space. However, this may be explained because optimizing a single linear layer may exhibit less inherent variance and covariance than multiple layers.

Overall, we have demonstrated the potential of linear policies on popular RL benchmarks. We showed that ES are effective optimizers for these policies compared to gradient-based methods. Additionally, we note that ES are simpler in design, have fewer hyperparameters, and are trivially parallelizable. Hence, ES can perform more environment interactions within the same time frame~\cite{salimans2017evolution}.
Moreover, evaluating linear policies is faster than evaluating one or sometimes several deep architectures, making the training much more expedient regarding wall-clock time. As the need for energy-efficient policy networks increases, our results warrant a closer look at ES for RL and training of simpler policies for tasks currently considered complex. 
For future work, we aim to extend our benchmark with more types of classical ES and strategies for multimodal optimization~\cite{preuss2015multimodal}. Additionally, it would be interesting to study the effect of the step-size adaptation methods in the presence of one or more hidden layers. Next, we will explore the potential of linear networks for other applications. Inspired by works such as~\cite{cuccu2018playing}, we will look at more complex Atari games to see if they can be solved by simple, energy-efficient means. 

\appendix

\section*{Setup}
For our experiments, we utilized three computing machines. The first machine was configured with an AMD 3950x processor, 16 CPU cores, 64 GB of RAM, and a GeForce RTX 3060 GPU. The second machine contained an Intel Core i9-13900K processor with 24 CPU cores and was equipped with 32 GB of RAM. The third machine was a Debian 12 server
running on two AMD EPYC 7662 64-core processors and 1\,000 GB RAM. The first two machines were used to obtain the results for the gradient-based algorithms, and the third was used to run the experiments with the ES, which do not require GPU acceleration. 

\section*{Hyperparameter Settings and Additional Results}

\begin{table}[!h]
\caption{Overview of the dimensionality of states and actions and the corresponding number of trainable weights for the linear neural network architecture.}
\centering
\begin{tabular}{l|rrr}
Environment &  Inputs &  Outputs & Weights \\
\hline
CartPole-v1 & 4 & 2 & 8 \\
Acrobot-v1 & 6 & 3 & 18 \\
Pendulum-v1 & 3 & 1 & 3 \\ 
\hline
LunarLander-v2 & 8 & 4 & 32 \\
BipedalWalker-v3 & 24 & 4 & 96 \\
\hline
Swimmer-v4 & 8 & 2 & 16 \\
Hopper-v4 & 11 & 3 & 33 \\
HalfCheetah-v4 & 17 & 6 & 102 \\
Walker2d-v4 & 17 & 6 & 102 \\
Ant-v4 & 27 & 8 & 216 \\
Humanoid-v4 & 376 & 17 & 6392 \\
\hline
Atari-v5 & 128 & $[4, 18]$ & $[512, 2304]$ \\

\hline
\end{tabular}
\label{tab:weights}
\end{table}

\begin{table}[!h]
\centering
\caption{Hyperparameters used in the ES configuration
    $\lambda_{\textsc{def}} = \min (128, \max (32, \frac{n}{2})), \lambda_{\textsc{cma}}$: see \protect\cite{hansen2001completely}.} 
\begin{tabular}{l|rrr}
 Environment & $\sigma^{(0)}$ & $\lambda$ \\ 
\hline
CartPole-v1 & 0.1 & 4 \\
Acrobot-v1 & 0.05 & 4 \\
Pendulum-v1 & 0.1 & $\lambda_{\textsc{def}}$ \\
\hline
LunarLander-v2 & 0.1 & $\lambda_{\textsc{def}}$ \\
BipedalWalker-v3 & 0.1 & $\lambda_{\textsc{def}}$\\
\hline
Swimmer-v4 & 0.1 & 4 \\
HalfCheetah-v4 & 0.05 & $\lambda_{\textsc{cma}}$ \\
Hopper-v4 & 0.05 & $\lambda_{\textsc{def}}$ \\
Walker2d-v4 & 0.05 & $\lambda_{\textsc{def}}$ \\
Ant-v4 & 0.05 & $\lambda_{\textsc{def}}$ \\
Humanoid-v4 & 0.01 & $\lambda_{\textsc{def}}$ \\ \hline 
Atari-v5 & 1.0 & $\lambda_{\textsc{def}}$ \\
\hline 
\end{tabular}

\label{tab:hyper}
\end{table}

\begin{table}[!h]
\centering
\caption{DQN architecture. In contrast to the original DQN, which utilizes convolutional layers for image processing, our approach is tailored for non-image data. This configuration is adapted from the CleanRL implementation ~\protect\cite{huang2022cleanrl}. }
\begin{tabular}{l|r}
Layer & Number of Nodes \\
\hline
Input Layer & Size of observation space \\
\hline
ReLU Hidden Layer 1 & 120 \\
\hline
ReLU Hidden Layer 2 & 84 \\
\hline
Output Layer & Size of action space \\ \hline
\end{tabular}
\label{tab:qnetwork_layers}
\end{table}

\begin{table}[!h]
\centering
\caption{DQN hyperparameters. This configuration is adapted from the CleanRL implementation ~\protect\cite{huang2022cleanrl}.}
\begin{tabular}{l|r}
Parameter & Value \\ \hline
Optimizer & Adam  \\ \hline 
Learning rate & $ 2.5 \cdot 10^{-4}$ \\ \hline
Discount factor ($\gamma$) & 0.99 \\ \hline
Replay buffer size & $ 1 \cdot 10^{4}$ \\ \hline
Target network update frequency (timesteps) & 500 \\ \hline
Batch size & 128 \\ \hline
Start-$\epsilon$ & 1 \\ \hline
End-$\epsilon$ & 0.05 \\ \hline
Fraction of timesteps to go from start-$\epsilon$ to end-$\epsilon$ & 0.5 \\ \hline
\end{tabular}
\label{tab:dqn_hyperparameters}
\end{table}

\begin{table}[!h]
\centering
\caption{PPO architecture. The PPO architecture employs the same actor-critic network structure as introduced in the original paper \protect\cite{schulman2017proximal}}.
\begin{tabular}{l|r|r}
Component & Layer & Number of Nodes \\
\hline
\multirow{4}{*}{Critic} & Input Layer & Observation space size \\
\cline{2-3}
& Tanh Hidden Layer 1 & 64 \\
\cline{2-3}
& Tanh Hidden Layer 2 & 64 \\
\cline{2-3}
& Output Layer & 1 \\
\hline
\multirow{4}{*}{Actor} & Input Layer & Observation space size \\
\cline{2-3}
& Tanh Hidden Layer 1 & 64 \\
\cline{2-3}
& Tanh Hidden Layer 2 & 64 \\
\cline{2-3}
& Output Layer & Action space size \\ \hline
\end{tabular}
\label{tab:ppo_layers}
\end{table}

\begin{table}[!h]
\centering
\caption{PPO Hyperparameters. Hyperparameters used for Classic Control and MuJoCo tasks, based on \protect\cite{schulman2017proximal}}
\begin{tabular}{l|r|r}
Hyperparameter & Classic Control & MuJoCo \\ \hline
Horizon (T) & 128 & 2048 \\ \hline
Optimizer & \multicolumn{2}{c}{Adam} \\ \hline
Clip coefficient ($\epsilon$) & \multicolumn{2}{c}{0.2} \\ \hline
Discount factor ($\gamma$) & \multicolumn{2}{c}{0.99} \\ \hline
GAE parameter ($\lambda$) & \multicolumn{2}{c}{0.95} \\ \hline
Mini batch size & 32 & 64 \\ \hline
\end{tabular}
\label{tab:ppo_hyperparameters_combined}
\end{table}



\begin{table}[!h]
\caption{SAC architecture: SAC employs the same actor-critic network structure as introduced in the original paper \protect\cite{haarnoja2018soft}}
\centering
\begin{tabular}{>{\raggedright}p{1.5cm}|>{\raggedright}p{2.5cm}|>{\raggedleft\arraybackslash}p{3cm}}
Component & Layer & Number of Nodes \\
\hline
\multirow{5}{2cm}{Actor} & Input Layer & Observation space size \\
\cline{2-3}
& ReLU Hidden Layer & 256 \\
\cline{2-3}
& ReLU Hidden Layer & 256 \\
\cline{2-3}
& Output Layer (Mean) & Size of action space \\
\cline{2-3}
& Output Layer (Log Std) & Size of action space \\
\hline
\multirow{4}{2cm}{Critic} & Input Layer & Observation space size + Action space size \\
\cline{2-3}
& ReLU Hidden Layer & 256 \\
\cline{2-3}
& ReLU Hidden Layer & 256 \\
\cline{2-3}
& Output Layer & 1 \\ \hline
\end{tabular}
\label{tab:sac_layers}
\end{table}

\begin{table}[!h]
\caption{SAC hyperparameters, based on \protect\cite{haarnoja2018soft}}
\centering
\begin{tabular}{l|r}
Parameter & Value \\ \hline
Optimizer & Adam  \\ \hline
Policy learning rate & $3 \times 10^{-4}$ \\ \hline
Q network learning rate & $1 \times 10^{-3}$ \\ \hline
Discount factor ($\gamma$) & 0.99 \\ \hline
Replay buffer size & $10^6$ \\ \hline
Mini batch size & 256 \\ \hline
Entropy target & - dimension(action space)  \\ \hline
Target smoothing coefficient ($\tau$) & 0.005 \\ \hline
Target update interval & 1 \\ \hline
\end{tabular}
\label{tab:sac_hyperparameters}
\end{table}

\begin{table}[!h]
\caption{ARS hyperparameters. We use the optimized hyperparameters for the MuJoCo environments as reported in the original study, and the default settings for the rest of the environments.}
\centering
\begin{tabular}{l|r|r|r|r}
Environment & $\alpha$ & $\nu$ & $N$ & $b$ \\
\hline
Swimmer-v4 & 0.02 & 0.01 & 1 & 1 \\
Hopper-v4 & 0.01 & 0.025 & 8 & 4 \\
HalfCheetah-v4 & 0.02 & 0.03 & 32 & 4 \\
Walker2d-v4 & 0.03 & 0.025 & 40 & 30 \\
Ant-v4 & 0.015 & 0.025 & 60 & 20 \\
Humanoid-v4 & 0.02 & 0.0075 & 230 & 230 \\
Other & 0.02 & 0.05 & 8 & 8  \\ \hline 
\end{tabular}
\label{tab:ars_hyperparameters}
\end{table}

\clearpage
\begin{figure*}[h]
    \centering
    \includegraphics[width=\linewidth]{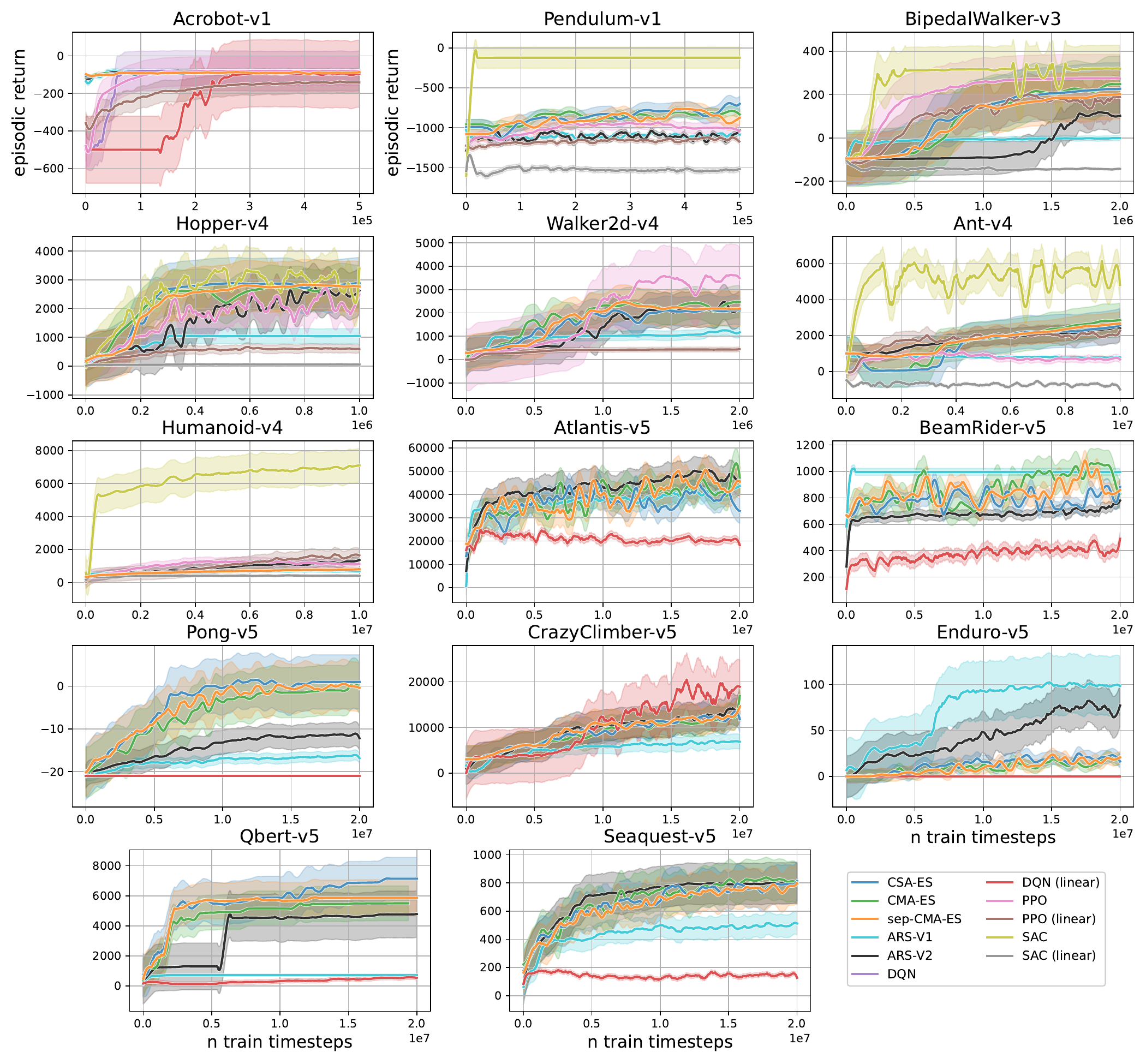}
    \caption{Training curves for the Classic, MuJoCo, and Atari environments. Episodic return (calculated using 5 test episodes) versus the number of training timesteps is shown. Each curve represents the median of 5 trial runs conducted with different random seeds; the shaded area denotes standard deviations. In the classic control environment Acrobat, linear ES and ARS solve the environment within few timesteps, exceeding the performance of gradient-based methods. In contrast, classic SAC excels in the Pendulum task and is the only method that achieves the maximum reward of 0. In the BipedalWalker environment, while classic PPO achieves the optimal reward of 300, both linear ES and linear PPO are not far behind in their performance. While ARS V2 outperforms ARS V1, it is still not as effective as the other methods. In the MuJoCo tasks, ES and ARS-v2 match the performance of the original DRL networks in Hopper, while the linear gradient-based methods struggle to learn a good policy. In Ant, Humanoid, and Walker2d, classic SAC emerges as the dominant method. The ES and ARS-v2 perform comparably to classic PPO, while linear PPO and linear SAC have difficulty finding a good policy. Interestingly, in the Ant environment, linear PPO succeeds in identifying an effective policy and even surpasses the performance of the larger PPO network. In the Atari environments Atlantis, BeamRider, Pong, Crazy Climber, Enduro, Qbert, and Seaquest, linear ES learns effective policies from the game's RAM as policy input. However, linear DQN fails to do the same, except for CrazyClimber, surpassing ES in performance. ARS-v1 outperforms all other methods in BeamRider and Enduro but performs not as well in Pong and Qbert.}
    \label{fig:combined}
\end{figure*}

\clearpage

\bibliographystyle{IEEEtran}
\bibliography{main}

\clearpage

\end{document}